\documentclass{article}
\usepackage{spconf,amsmath,graphicx}
\usepackage{MnSymbol}
\usepackage{bbding}
\usepackage{amsfonts}
\usepackage{times}
\usepackage{url}
\usepackage{hyperref}
\usepackage{multirow}
\usepackage{hhline}
\usepackage{makecell}
\usepackage{arydshln}

\newcommand{\newcite}[2]{#1 et al.~\cite{#2}}
\newcommand{\parjump}{\vspace{+0.2em}}
\newcommand{\newparagraph}[1]{\par \parjump \noindent \textbf{#1}~~}

\title{Pro-HAN: A Heterogeneous Graph Attention Network for Profile-Based Spoken Language Understanding}
\name{Dechuan Teng$^{1}$ \qquad Chunlin Lu$^{2}$ \qquad Xiao Xu$^{1}$ \qquad Wanxiang Che$^{1\dagger}$ \qquad Libo Qin$^{2\dagger}$
\thanks{$^{\dagger}$Corresponding author.}
}
\address{$^{1}$Research Center for Social Computing and Information Retrieval, Harbin Institute of Technology, China \\
$^{2}$School of Computer Science and Engineering, Central South University, China}

\begin{document}
%
\maketitle
\begin{abstract}
    Recently, Profile-based Spoken Language Understanding (SLU) has gained increasing attention, which aims to incorporate various types of supplementary profile information (i.e., \textit{Knowledge Graph}, \textit{User Profile}, \textit{Context Awareness}) to eliminate the prevalent ambiguities in user utterances.
    However, existing approaches can only separately model different profile information, without considering their interrelationships or excluding irrelevant and conflicting information within them.
    To address the above issues, we introduce a \textbf{H}eterogeneous Graph \textbf{A}ttention \textbf{N}etwork to perform reasoning across multiple \textbf{\textsc{Pro}}file information, called \textbf{\textsc{Pro}-HAN}.
    Specifically, we design three types of edges, denoted as \textit{intra-\textsc{Pro}}, \textit{inter-\textsc{Pro}}, and \textit{utterance-\textsc{Pro}}, to capture interrelationships among multiple \textsc{Pro}s.
    We establish a new state-of-the-art on the ProSLU dataset, with an improvement of approximately 8\% across all three metrics.
    Further analysis experiments also confirm the effectiveness of our method in modeling multi-source profile information.

\end{abstract}
    
\section{Introduction}
\label{sec:intro}

\begin{figure*} [t]
	\centering
	\includegraphics[width=0.7\textwidth]{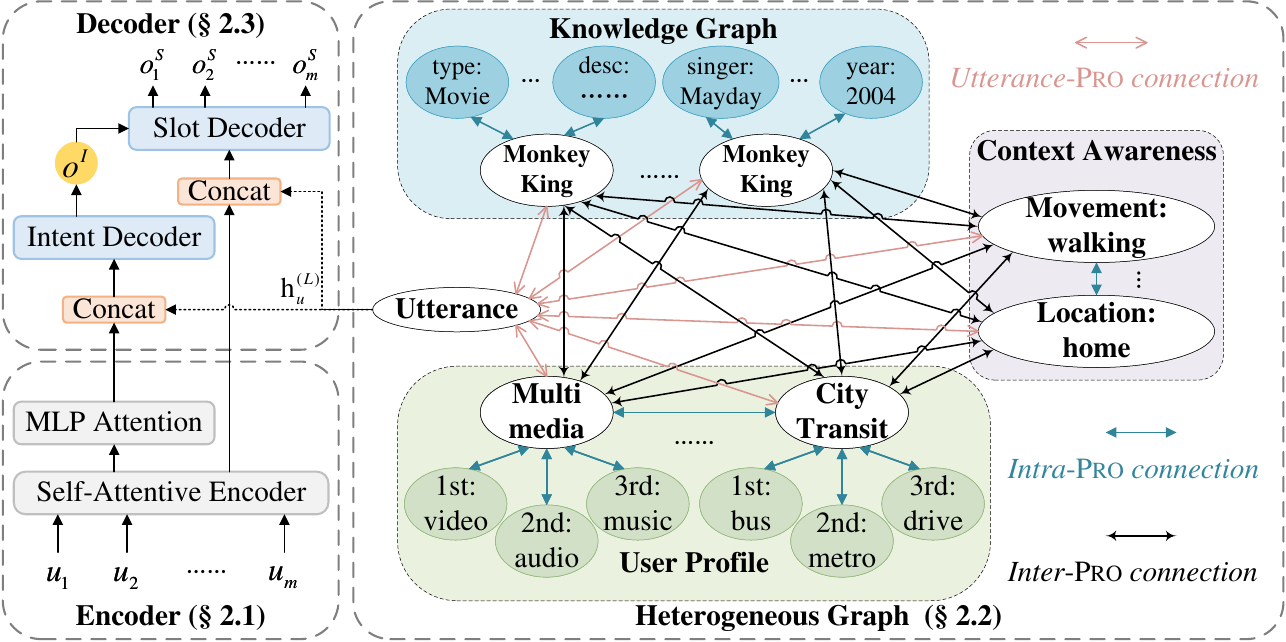}
	\caption{The illustration of \textsc{Pro}-HAN. The initial representations of graph nodes are introduced in Section~\ref{sec:encoder}.}
	\label{fig:model_framework}
\end{figure*}

As an important module in task-oriented dialogue systems, Spoken Language Understanding (SLU) assists in subsequent modules (i.e., dialogue management and response generation) by recognizing the intent and related slots in a user utterance~\cite{tur2011spoken,DBLP:journals/pieee/YoungGTW13,DBLP:conf/ijcai/QinXC021}.
The traditional SLU systems assume that users' demands can be fully determined based solely on user utterances.
However, in real-world scenarios, the presence of semantic ambiguity within user utterances makes it challenging to precisely ascertain their demands.
For example, the subject ``\textit{Half a Lifelong Romance}" mentioned in the utterance in Table~\ref{tab:complex_example} can be either a movie, a music, or a novel.

\begin{table}[t]
    \centering
    \resizebox{0.5\textwidth}{!}{
        \begin{tabular}{l|l}
            \hline \hline \multicolumn{2}{c}{ \textbf{Input} } \\
            \hline \textbf{Text} & \textit{Playing Eileen Chang's ``Half a Lifelong Romance" on TV} \\
            \hline \textbf{KG} & \makecell[l]{ \texttt{Mention} "\textit{Half a Lifelong Romance}": \textbf{10} subjects  \\ \hdashline type: \textit{Creative}, singer: \textit{Leon Lai}, tag: \textit{music}, ... \\ type: \textit{Movie}, director: \textit{Ann Hui},  starring: \textit{Jacklyn Wu}, ... \\ type: \textit{Creative}, author: \textit{Eileen Chang}, tag: \textit{literature}, ... } \\
            \hline \textbf{UP} & \makecell[l]{ Preference for [music, video \& audiobook]: [\textit{0.1,0.7,0.2}], \\ Preference for [metro, bus \& drive]: [\textit{0.3,0.6,0.1}], ... } \\
            \hline \textbf{CA} & Movement State: \textit{walking}, Geographic Location: \textit{home}, ... \\
            \hline \multicolumn{2}{c}{ \textbf{Output} } \\
            \hline \textbf{Intent} & \makecell[l]{\texttt{PlayVoice}} \\
            \hline \textbf{Slot} & \makecell[l]{artist: \texttt{Eileen Chang},~~deviceType: \texttt{TV}, \\ voiceName: \texttt{Half a Lifelong Romance}} \\
            \hline
        \end{tabular}
    }
    \caption{An example from the \textsc{ProSLU} dataset.}
	\label{tab:complex_example}
\end{table}

To overcome the above limitations, \newcite{Xu}{xu2022text} introduced a new task and benchmark, Profile-based Spoken Language Understanding (\textsc{ProSLU}), which provides three types of supporting information for each user utterance: \textit{Knowledge Graph} (KG), \textit{User Profile} (UP), and \textit{Context Awareness} (CA), referred to as \textsc{Pro}s.
In addition, \newcite{Xu}{xu2022text} proposed a GSM++ framework that statically and separately encodes each \textsc{Pro}, then applies an attention mechanism to integrate them, and finally injects the fused profile features into a general SLU model to identify users' actual intents and slots.

Despite achieving the promising performance, the baseline model has several limitations:
(1) Firstly, a large amount of irrelevant information exists in \textsc{Pro}s.
In particular, KG contains numerous homonymous subjects (e.g., \textit{Half a Lifelong Romance}) and their rich attributes.
GSM++ represented the entire KG by averaging the encodings of all subjects, where each subject and its attributes are flattened into a sequence, making it hard to distinguish the correct subject from the homonymous ones;
(2) Secondly, conflicts may also arise among \textsc{Pro}s.
For instance, \textit{``Half a Lifelong Romance"} authored by \textit{Eileen Chang} is a novel, but the user prefers videos, in which case the user's multimedia preference could not be satisfied.
However, separately modeling different \textsc{Pro}s fails to effectively reason across multiple \textsc{Pro}s;
(3) Lastly, the static modeling of \textsc{Pro}s without considering the utterance makes it difficult to adaptively extract and model valuable profile information based on the given user utterance.

To solve the aforementioned challenges, we construct a \textbf{H}eterogeneous Graph \textbf{A}ttention \textbf{N}etwork over multiple \textbf{\textsc{Pro}}file information (\textsc{Pro}-HAN).
The core of \textsc{Pro}-HAN is the profile information interaction based on three types of heterogeneous graph connections.
Specifically, to address the first challenge, we introduce the \textit{Intra-\textsc{Pro} connection} to depict the inherent structural relationship within each supporting information, laying the foundation for comprehensive message aggregation.
For the second challenge, the \textit{Inter-\textsc{Pro} connection} between multiple \textsc{Pro}s is designed to establish the information flow across them.
For the third challenge, the \textit{Utterance-\textsc{Pro} connection} allows the model to adaptively integrate useful profile information based on the given user utterance.

Experimental results on the \textsc{ProSLU} dataset show \textsc{Pro}-HAN outperforms previous SOTA method by 8\% on all metrics.
Besides, further analysis experiments demonstrate that \textsc{Pro}-HAN effectively utilizes multiple \textsc{Pro}s to disambiguate utterances based on the proposed heterogeneous connections.
All codes for this work are publicly available at \href{https://github.com/AaronTengDeChuan/PRO-HAN}{https://github.com/AaronTengDeChuan/PRO-HAN}.

\section{Approach}
\label{sec:approach}

This section introduces the architecture of \textbf{H}eterogeneous Graph \textbf{A}ttention \textbf{N}etwork for \textbf{\textsc{Pro}}file-based SLU (\textsc{Pro}-HAN), which is illustrated in Figure~\ref{fig:model_framework}.
 \textsc{Pro}-HAN consists of three components: an encoder to represent utterance and multiple \textsc{Pro}s (\S\ref{sec:encoder}), a heterogeneous graph attention network to model the interrelationships among multiple Pros (\S\ref{sec:graph_network}), and a SLU decoder to predict intent and slots (\S\ref{sec:decoder}).

\subsection{Encoder}
\label{sec:encoder}

\newparagraph{UP and CA Representation}
\textbf{U}ser \textbf{P}rofile (UP) contains user preferences for different categories, such as multimedia and city transit.
Each option of each preference is represented in the form of a triplet $\langle \text{option}, \text{order}, \text{category} \rangle$.
For example, $\langle \texttt{music}, \texttt{third}, \texttt{multimedia} \rangle$ indicates that music is the user's third-favorite multimedia genre.
\textbf{C}ontext \textbf{A}wareness (CA) denotes the user states, such as movement state and geographic location.
Each category in CA is also represented as a triplet $\langle \text{state}, \text{category}, \text{context awareness} \rangle$, such as $\langle \texttt{walking}, \texttt{movement state}, \texttt{context awareness} \rangle$.

We adopt the approach proposed by~\newcite{Qin}{DBLP:conf/aaai/0001LYWC23} for representing triplets, where we first apply a word embedding function $\phi^{emb}$ to obtain the embedding of each element in a triplet, and then sum them up.
Finally, we get the representations for UP and CA:
\begin{equation}
	\begin{aligned}
		\mathcal{P}_{\text{up}} &= \{p^{1}_{\text{up},1},\ldots,p^{1}_{\text{up},m_1},\ldots,p^{N_{\text{up}}}_{\text{up},1},\ldots,p^{N_{\text{up}}}_{\text{up},m_{N_{\text{up}}}}\},
		\\
		\mathcal{P}_{\text{ca}} &= \{p^{1}_{\text{ca}},\ldots,p^{N_{\text{ca}}}_{\text{ca}}\} \in \mathbb{R}^{N_{\text{ca}}\times d_e},
	\end{aligned}
	\label{eq:up_ca_reps}
\end{equation}
where $N_{up}$ and $N_{ca}$ are the number of categories in UP and CA, $m_{i}$ is the number of options in the $i$-th category of UP, and $d_e$ is the word embedding dimension.

\newparagraph{Text Encoder}
Following~\newcite{Qin}{qin2019stack}, we adopt a self-attentive encoder to obtain text representations.
Given the input text $ \mathbf{x}=\left\{ x_1,x_2,\ldots,x_T \right\} $ with $T$ tokens, a BiLSTM~\cite{hochreiter1997long} and a self-attention layer~\cite{DBLP:conf/nips/VaswaniSPUJGKP17} are combined to capture the sequential and contextual information, denoted as $\mathbf{E}=\left\{ \mathbf{e}_1,\mathbf{e}_2,\ldots,\mathbf{e}_T \right\} = \operatorname{SAEncoder}\left(\mathbf{x}\right) $.
Then, the whole text representation $\mathbf{g}=\operatorname{MLPAttn}\left(\mathbf{E}\right)$ is obtained by applying an MLP attention module~\cite{zhong2018global,zhang2018neural} over $\mathbf{E}$.
Hence, a text encoder can be formulated as $\mathbf{E},\mathbf{g}=\operatorname{TextEncoder}\left( \mathbf{x} \right)$.

\newparagraph{KG Representation}
\textbf{K}nowledge \textbf{G}raph (KG) contains many subjects mentioned in user utterances and their rich attributes, as shown in Table~\ref{tab:complex_example}.
Following~\newcite{Eric}{eric2017key}, each attribute is represented as a triplet $\langle \text{entity}, \text{attribute}, \text{subject} \rangle$.
We then use a text encoder to separately encode three elements in each triplet, and then sum them up to obtain the representation of each attribute.
Finally, we can get the KG representations:
\begin{equation}
		\mathcal{P}_{\text{kg}} = \{p^{1}_{\text{kg},1},\ldots,p^{1}_{\text{kg},n_1},\ldots,p^{N_{\text{kg}}}_{\text{kg},1},\ldots,p^{N_{\text{kg}}}_{\text{kg},n_{N_{\text{kg}}}}\},
	\label{eq:kg_reps}
\end{equation}
where $N_{kg}$ is the number of subjects in KG, and $n_{i}$ is the number of attributes of the $i$-th subject.

\newparagraph{Utterance Representation}
For the user utterance $\mathbf{u}$ with $m$ tokens, we use another text encoder to obtain the utterance representation $\mathbf{h}\in \mathbb{R}^{d_u}$ and token representations $\mathbf{U}\in \mathbb{R}^{m\times d_u}$.

\subsection{Heterogeneous Graph Attention Network}
\label{sec:graph_network}

In this section, we first construct a heterogeneous graph $\mathcal{G}=\{\mathcal{V},\mathcal{E}\}$, and then introduces
the message aggregation process.

\newparagraph{Node Building}
Graph Nodes $\mathcal{V}$ consists of an utterance node and all triplet nodes from KG, CA, and UP.
Besides,we also add some extra nodes to facilitate the aggregation of graph features.
We add a global node connected to all corresponding attribute nodes for each subject in KG, and add a global node connected to its option nodes for each category in UP.
For CA, each state node can be regarded as a global node.

We initialize the utterance node and triplet node features with utterance representation $\mathbf{h}$ and corresponding triplet representations.
After initialization, four different node linear transformations are applied to distinguish different nodes.

\newparagraph{Edge Building}
The following three types of edges constitute the graph connections $\mathcal{E}$.

\begin{itemize}
\item \textbf{Intra-\textsc{Pro} Connection} is used for modeling internal information within each \textsc{Pro}.
All global nodes with different subject names in KG, and all global nodes in UP and CA are interconnected, respectively.
\item \textbf{Inter-\textsc{Pro} Connection} incorporates cross-\textsc{Pro}s information, where three types of global nodes are linked to each other.
\item \textbf{Utterance-\textsc{Pro} Connection} is established between the utterance node and all global nodes to adaptively integrate useful clues based on the given utterance.
\end{itemize}

\newparagraph{Heterogeneous Message Aggregation}
We apply $L$ graph attention layers~\cite{brody2021attentive} to implement message passing and aggregation over graph nodes $\mathcal{V}$, which can be formulated as:
\begin{equation}
	\begin{aligned}
		h_i^{(l+1)} &= \sum_{r \in \mathcal{R}} f_r( \sum_{j \in \mathcal{N}(i,r)} \alpha_{i j}^{(l)} W_{\text {right }}^{(l)} h_j^{(l)}), \\
		\alpha_{i j}^{(l)} & =\operatorname{softmax}_{\mathrm{i}}\left(e_{i j}^{(l)}\right), \\
		e_{i j}^{(l)} & =\vec{a}^{T^{(l)}} \operatorname{LeakyReLU}\left(W_{\text {left }}^{(l)} h_i^{(l)}+W_{\text {right }}^{(l)} h_j^{(l)}\right),
	\end{aligned}
\end{equation}
where $h_i^{(l)}$ is the hidden states of node $i$ in layer $l$, $\mathcal{N}(i,r)$ is the neighbors of node $i$ with edge type $r\in\mathcal{R}$, $\alpha_{i j}^{(l)}$ is the attention score between node $i$ and node $j$, and $f_r$ is a linear transformation function for edge type $r$.

After $L$ layers of message aggregation, we obtain the final node representations $\mathbf{H}^{(L)}=\left\{ \mathbf{h}_1^{(L)},\mathbf{h}_2^{(L)},\ldots,\mathbf{h}_{N}^{(L)} \right\}$, where $N$ is the number of nodes in the graph and $\mathbf{h}_u^{(L)}$ is the representation of the utterance node.

\subsection{SLU Decoder}
\label{sec:decoder}
The SLU decoder consists of two parts: intent decoder and slot decoder.
Through the \textit{Utterance-\textsc{Pro} Connection}, the final utterance node representation $\mathbf{h}_u^{(L)}$ has already integrated the information of all \textsc{Pro}s, which can directly be used to enhance intent detection and slot filling.

\newparagraph{Intent Decoder}
The utterance representation $\mathbf{h}$ and utterance node representation $\mathbf{h}_u^{(L)}$ are fused to predict the intent:
\begin{equation}
	\begin{aligned}
		\mathbf{y}^{\mathrm{I}} & =\operatorname{softmax}\left(\mathbf{W}_{\mathbf{I}} \left[ \mathbf{h} \parallel \mathbf{h}_u^{(L)} \right]\right), \\
	\end{aligned}
\end{equation}
where $\parallel$ denotes the concatenation operation.
The predicted intent is obtained by $o^{\mathrm{I}} =\arg \max \left(\mathbf{y}^{\mathrm{I}}\right)$.

\newparagraph{Slot Decoder}
A unidirectional LSTM with the same intent-guided mechanism as~\newcite{Qin}{qin2019stack} is used to decode the slot label sequence $\{o^{\mathrm{S}}_1,\ldots,o^{\mathrm{S}}_m\}$.
Specifically, the intent embedding $\phi^{int}\left(o^{\mathrm{I}}\right)$ and utterance node representation $\mathbf{h}_u^{(L)}$ are first concatenated to enhance token representations.
Then, the hidden states $\mathbf{h}_{t}$ of the slot filling decoder are computed as:
\begin{equation}
	\begin{aligned}
		\mathbf{h}_{t}^{\mathrm{S}} &=\operatorname{LSTM}\left(\mathbf{h}_{t-1}^{\mathrm{S}}, \mathbf{s}_{t} \parallel \phi^{int}\left(o^{\mathrm{I}}\right) \parallel \mathbf{h}_u^{(L)}\right), \\
		\mathbf{y}_{t}^{\mathrm{S}} &=\operatorname{softmax}\left(\mathbf{W}_{\mathbf{S}} \mathbf{h}_{t}^{\mathrm{S}}\right), \\
	\end{aligned}
\end{equation}
where $\mathbf{s}_{t}$ is the concatenation of the encoder hidden states $\mathbf{u}_{t}$ and the previous slot label embedding $\phi^{slot}\left(o_{t-1}^{\mathrm{S}}\right)$.
\section{Experiments}
\label{sec:experiments}

\begin{table}[t]
    \centering
    \resizebox{0.49\textwidth}{!}{
        \begin{tabular}{l|ccc}
            \hline \hline
            \textbf{Model} & \textbf{Slot ($ F_1 $)} & \textbf{Intent ($ Acc $)} & \textbf{Overall ($ Acc $)} \\ \cline{2-4}
            \hline
            Slot-Gated$^{\ddag}$~\cite{goo2018slot} & 74.18 & 83.24 & 69.11 \\
            Bi-Model$^{\ddag}$~\cite{wang2018bi} & 77.76 & 82.30 & 73.45 \\
            SF-ID$^{\ddag}$~\cite{haihong2019novel} & 73.70 & 83.24 & 68.36 \\
            Stack-Propagation$^{\ddag}$~\cite{qin2019stack} & 81.08 & 83.99 & 78.91 \\
            GSM++$^{\ddag}$~\cite{xu2022text} & 83.27 & 85.31 & 79.10 \\
            \hdashline
            \textsc{Pro}-HAN & \textbf{91.62} & \textbf{94.16} & \textbf{87.01} \\
            \hline
            GPT-3.5 (0-shot) & 35.77 & 39.74 & 31.83 \\
            GPT-3.5 (14-shot) & 38.60 & 41.43 & 34.27 \\
            \hline \hline
        \end{tabular}
    }
    \caption{Main results on the \textsc{Pro}SLU dataset.
    \ddag~indicates the results are quoted from~\newcite{Xu}{xu2022text}.}
	\label{tab:Main resuts}
\end{table}

\subsection{Dataset \& Metrics}
We conduct experiments on the \textsc{ProSLU} dataset~\cite{xu2022text}, which contains 4,196 training, 522 validation, and 531 test samples, covering 14 intents and 49 slots.
Following~\newcite{Goo}{goo2018slot} and~\newcite{Qin}{qin2019stack}, we adopt F1 score, intent accuracy and overall accuracy as the metrics for \textsc{ProSLU}.

\subsection{Main Results}
Table~\ref{tab:Main resuts} shows the results of the proposed method and strong baselines on the \textsc{ProSLU} dataset.
We observe that \textsc{Pro}-HAN surpasses the previous SOTA model (GSM++~\cite{xu2022text}) by approximately 8\% in terms of all three metrics, and thereby establishes a new state-of-the-art.
This indicates that our heterogeneous graph attention network can better extract useful profile information to mitigate ambiguity in user utterances compared to the Multi-Level Knowledge Adapter in GSM++.

\subsection{Analysis}

\newparagraph{Intra-\textsc{Pro} Connection Captures the Internal Structural Information}
We only remove the \textit{intra-\textsc{Pro} connection} from the set of graph edges $\mathcal{E}$ and the results is presented in Table~\ref{tab:Ablation study}~(\textit{w/o Intra-\textsc{Pro}}).
About 3\% performance drop is observed in all metrics, indicating that the \textit{intra-\textsc{Pro} connection} is effective in capturing the structural information within each type of profile information.

\newparagraph{Inter-\textsc{Pro} Connection Facilitates the Reasoning across Multiple \textsc{Pro}s}
We further investigate the impacts of the interaction among different \textsc{Pro}s by removing the \textit{inter-\textsc{Pro} connection}.
Table~\ref{tab:Ablation study}~(\textit{w/o Inter-\textsc{Pro}}) presents a significant decrease of more than 5\% across all three metrics, illustrating that the information reasoning among multiple \textsc{Pro}s facilitates accurate recognition of user demands.

\newparagraph{Utterance-\textsc{Pro} Connection Enables Adaptive Information Extraction}
The most severe performance degradation (up to 14\%) occurs when breaking the connection between utterance and \textsc{Pro}s, as shown in Table~\ref{tab:Ablation study}~(\textit{w/o Utterance-\textsc{Pro}}).
This is attributed to the challenge posed by separately modeling both, making it difficult for the model to dynamically extract cues from \textsc{Pro}s.

\newparagraph{Heterogeneous GAT vs. Homogeneous GAT}
We replace the heterogeneous GAT with a homogeneous GAT, which means that all edges in the graph network are of the same type.
As shown in Table~\ref{tab:Ablation study}~(\textit{Homogeneous GAT}), the overall accuracy decreased from 87\% to 79\% after the replacement.
This suggests that the homogeneous GAT is not as proficient as its heterogeneous counterpart in distinguishing among diverse sources of information, resulting in the confusion when integrating information interactions.
Consequently, it becomes challenging to retrieve relevant cues across \textsc{Pro}s.

\begin{table}[t]
    \centering
    \resizebox{0.49\textwidth}{!}{
        \begin{tabular}{l|ccc}
            \hline
            \textbf{Model} & \textbf{Slot ($ F_1 $)} & \textbf{Intent ($ Acc $)} & \textbf{Overall ($ Acc $)} \\ \cline{2-4}
            \hline
            \textsc{Pro}-HAN & \textbf{91.62} & \textbf{94.16} & \textbf{87.01} \\
            \hdashline
            w/o Intra-\textsc{Pro} & 88.62 & 91.15 & 83.62 \\
            w/o Inter-\textsc{Pro} & 86.94 & 88.51 & 81.92 \\
            w/o Utterance-\textsc{Pro} & 78.83 & 79.85 & 74.39 \\
            Homogeneous GAT & 81.35 & 84.18 & 79.10 \\
            \hline
        \end{tabular}
    }
    \caption{Ablation study of \textsc{Pro}-HAN.}
	\label{tab:Ablation study}
\end{table}

\newparagraph{Investigation on GPT-3.5}
Furthermore, we also evaluate the zero-shot and few-shot performance of GPT-3.5 on the \textsc{ProSLU} dataset.
In the few-shot setting, the prompt input to GPT-3.5 contains four parts: task description and schema, 14 demonstrations corresponding to each intent, regulations, and the test input.
From Table~\ref{tab:Main resuts}, we can see that GPT-3.5 performs poorly on the \textsc{ProSLU} dataset, indicating that GPT-3.5 fails to distinguish the clues that can determine user demands from the provided profile information.

\newparagraph{Case Study}
Thanks to the \textit{intra-\textsc{Pro}} and \textit{utterance-\textsc{Pro}} connections, our method can more accurately distinguish the entities in KG that are related to the user utterance.
For example, GSM++ predicts the intent of the utterance in Table~\ref{tab:complex_example} as \texttt{PlayVideo}, which is consistent with the user's multimedia preference, indicating that it did not find the correct subject in KG.
In contrast, our method can identify that ``\textit{Half a Lifelong Romance}" is a novel from KG, based on the author ``\textit{Eileen Chang}" mentioned in the utterance, and then obtain the correct intent ``\texttt{PlayVoice}".
Moreover, the \textit{inter-\textsc{Pro} connection} makes it possible to comprehensively consider multiple \textsc{Pro}s, allowing the model to selectively ignore conflicting information, i.e., the multimedia preference in UP in this case.

\section{Related Work}
\label{sec:related_work}

Dominant work in SLU literature considered interaction between intent detection and slot filling.
Initially, the two subtasks were implicitly connected by multi-task learning~\cite{zhang2016joint,hakkani2016multi,liu2016attention}.
Subsequently, an intent-guided mechanisms was proposed to guide the slot filling~\cite{goo2018slot,li2018self,qin2019stack,teng2021injecting}.
Another series of work explored establishing bidirectional connections between the two tasks~\cite{wang2018bi,haihong2019novel,zhang2019joint,liu2019cm,zhang2020graph,qin2021cointeractive}.
To make SLU models applicable to real-world scenarios, \newcite{Xu}{xu2022text} recently introduced additional profile information to assist in understanding user utterances, where three types of profile information are separately modeled in a static manner.
Compared to the static manner, we focus on how to explore heterogeneous graph network to adaptively and flexibly integrate information from different sources.

\section{Conclusion}
\label{sec:conclusion}

In this paper, we propose a heterogeneous graph attention network for Profile-based SLU, which enables the model to better eliminate ambiguity in user utterances by reasoning across multiple profile information (\textsc{Pro}s).
The significant performance improvement validates the advantages of heterogeneous graph in modeling multi-source information with complex structures.
In addition, we explore large language models (LLMs) for \textsc{ProSLU}, and results reveal that \textsc{ProSLU} is also challenging for LLMs.
\section{Acknowledgements}
\label{sec:acknowledgements}

We gratefully acknowledge the support of the National Natural Science Foundation of China (NSFC) via grant 62306342, 62236004 and 62206078, and the support of Du Xiaoman (Beijing) Science Technology Co., Ltd. This work was also sponsored by CCF-Baidu Open Fund.

\clearpage

\bibliographystyle{IEEEbib}
\bibliography{refs_SimBiber}

\end{document}